\documentclass[11pt]{article}

\usepackage[preprint]{acl}

\usepackage{times}
\usepackage{latexsym}
\usepackage{amsmath}
\usepackage{float}
\usepackage{enumitem}
\usepackage{cuted}
\usepackage{tikz}
\usepackage{pgfplots}
\pgfplotsset{compat=1.18}
\usetikzlibrary{positioning, arrows.meta, backgrounds, calc, shapes.geometric, fit}

\usepackage[T1]{fontenc}

\usepackage[utf8]{inputenc}

\usepackage{microtype}

\usepackage{inconsolata}
\usepackage{subcaption}

\usepackage{graphicx}
\usepackage{rotating}
\usepackage{booktabs}
\usepackage{multirow}
\usepackage[
    detect-all=true,
    detect-family=true,
    group-digits=false,
    separate-uncertainty=true
]{siunitx}
\renewrobustcmd{\bfseries}{\fontseries{b}\selectfont}
\renewrobustcmd{\boldmath}{}
\newrobustcmd{\B}{\bfseries}

\title{How Language Models Fail: Token-Level Signatures of Committed and Persistent Reasoning Failures}

\author{
 \textbf{Tanvi Thoria\textsuperscript{1}},
 \textbf{Kiana Jafari\textsuperscript{2}},
 \textbf{Marc R. Schlichting\textsuperscript{2}},
 \textbf{Mykel J. Kochenderfer\textsuperscript{1,2}},
\\
\\
 \textsuperscript{1}Department of Computer Science, Stanford University\\
 \textsuperscript{2}Department of Aeronautics and Astronautics, Stanford University,
\\
 \small{
   \textbf{Correspondence:} \href{mailto:mykel@stanford.edu}{mykel@stanford.edu}
 }
}

\begin{document}
\maketitle

\begin{abstract}
Failures in language model reasoning emerge through distinct processes that leave identifiable signatures in the reasoning trace. We characterize these failures using token-level uncertainty signals, finding they arise through two empirically distinguishable processes. The first is committed failure, in which a model locks onto an incorrect reasoning path early in its trace. A central diagnostic signature is the commitment point, beyond which considering additional tokens hurt rather than help failure detection. In the second, persistent uncertainty, uncertainty instead accumulates throughout, and the full trace is needed to best distinguish failing from successful completions. These signatures reproduce across 23 model-dataset configurations, with the framework's falsifiable predictions holding in 20 of 23 cases, well above chance across both failure modes. Finally, we demonstrate our failure mode framework has direct implications for self-consistency, identifying when uncertainty signals complement it and when it can be selectively skipped. These results offer a foundation for understanding when LLM reasoning failures become detectable and for adapting detection strategies accordingly. 
\end{abstract}

\section{Introduction}
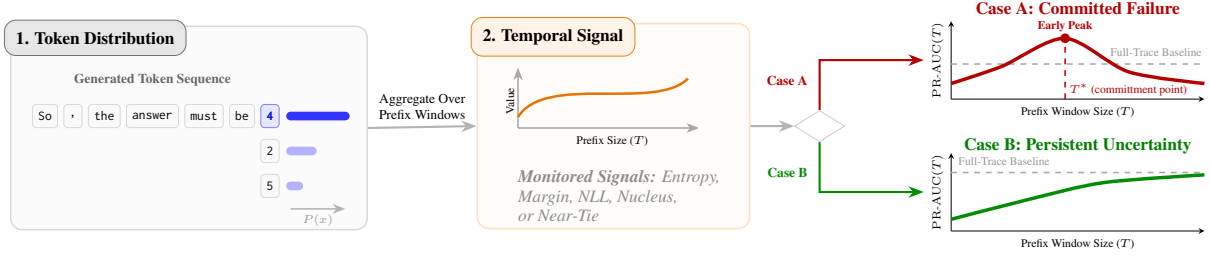
\begin{figure*}[t] 
    \centering
    \resizebox{\linewidth}{!}{%
    \begin{tikzpicture}[
        >=Stealth,
        font=\rmfamily,
        box/.style={draw=gray!30, thick, rounded corners=6pt, fill=white, inner xsep=4mm,inner ysep=0pt,text height=30mm,text depth=7mm,yshift=4mm},
        title/.style={font=\small\bfseries, fill=white, inner sep=6pt, rounded corners=6pt, xshift=-3pt, yshift=3pt,draw=black!10!white, thin},
        token/.style={draw=gray!30, fill=gray!5, rounded corners=2pt, inner sep=3pt, font=\ttfamily\scriptsize, minimum height=12pt},
        selected_token/.style={draw=blue!50, fill=blue!10, rounded corners=2pt, inner sep=3pt, minimum height=12pt, font=\ttfamily\scriptsize\bfseries, text=blue!90!black},
        arr/.style={->, thick, draw=gray!60, line width=1.1pt},
        axis/.style={thick, ->, draw=gray!70}
    ]

    \begin{scope}[local bounding box=token_stage]
        \node[token] (t1) {So};
        \node[token, right=1mm of t1] (t2) {,};
        \node[token, right=1mm of t2] (t3) {the};
        \node[token, right=1mm of t3] (t4) {answer};
        \node[token, right=1mm of t4] (t5) {must};
        \node[token, right=1mm of t5] (t6) {be};
        \node[selected_token, right=1mm of t6] (t7) {4};
        
        \node[above=2mm of t4, font=\scriptsize\bfseries, text=gray!80!black] {Generated Token Sequence};
        
        \node[token, below=2mm of t7] (v1) {2};
        \node[token, below=2mm of v1] (v2) {5};
        
        \draw[blue!80, line width=4.5pt, line cap=round] ($(t7.east)+(2mm,0)$) -- ++(1.0, 0);
        \draw[blue!30, line width=4.5pt, line cap=round] ($(v1.east)+(2mm,0)$) -- ++(0.4, 0);
        \draw[blue!30, line width=4.5pt, line cap=round] ($(v2.east)+(2mm,0)$) -- ++(0.15, 0);
        
        \draw[thick, ->, draw=gray!60] ($(v2.south east)+(2mm,-2mm)$) -- ++(1.0,0) 
            node[below, midway, yshift=1pt, font=\tiny, text=gray!80] {$P(x)$};
    \end{scope}
    
    \begin{scope}[on background layer]
        \node[fit=(token_stage), box, fill=gray!3] (box_stage1) {};
        \node[title, anchor=north west, fill=black!10!white, draw=black!10!gray, thin] at (box_stage1.north west) {1. Token Distribution};
    \end{scope}

    \path let \p1 = ($(box_stage1.north)-(box_stage1.south)$) in \pgfextra{\xdef\boxheight{\y1}};

    \begin{scope}[local bounding box=signal_stage]
        \node[right=2.6cm of box_stage1.east, yshift=0mm] (sig_anchor) {};
        \draw[axis] (sig_anchor.center) -- ++(3.3,0) node[below, xshift=-45pt] {\tiny Prefix Size ($T$)};
        \draw[axis] (sig_anchor.center) -- ++(0,1.1) node[rotate=90, left=4pt, pos=0.8] {\tiny Value};
        
        \draw[thick, orange!90!black, line width=1.3pt] ($(sig_anchor.center)+(0,0.2)$) .. controls ++(0.5,0.8) and ++(-0.6,-0.6) .. ($(sig_anchor.center)+(3.1,0.9)$);
        
        \node[anchor=north west, font=\footnotesize\itshape, text=gray!90, align=left, yshift=-5mm] (signal_list) at (sig_anchor.south west) {
            \textbf{Monitored Signals:} Entropy,\\ Margin, NLL, Nucleus,\\ or Near-Tie
        };
    \end{scope}

    \begin{scope}[on background layer]
        \node[fit=(signal_stage) (signal_list), box, fill=orange!2, draw=orange!25, 
              minimum height=\boxheight] (box_stage2) {};
        \node[title, anchor=north west, fill=orange!10!white, draw=orange!80!gray] at (box_stage2.north west) {2. Temporal Signal};
    \end{scope}

    \draw[arr] (box_stage1.east) -- (box_stage2.west) 
        node[midway, above, align=center, font=\scriptsize] {Aggregate Over\\Prefix Windows};

    \node[draw=gray!40, fill=white, diamond, aspect=1.6, inner sep=5pt, right=0.8cm of box_stage2] (branch) {};
    \draw[arr] (box_stage2.east) -- (branch.west);

    \node[right=1.8cm of branch, yshift=1.2cm] (plot1_anchor) {};
    \node[right=1.8cm of branch, yshift=-1.2cm] (plot2_anchor) {};

    \begin{axis}[
        name=plot1,
        at={(plot1_anchor)}, anchor=west,
        width=6.2cm, height=3cm,
        title={\textbf{Case A: Committed Failure}},
        title style={font=\small, color=red!70!black, yshift=-6pt},
        xlabel={Prefix Window Size ($T$)},
        ylabel={$\mathrm{PR\text{-}AUC}(T)$},
        xlabel style={font=\tiny, yshift=2pt},
        ylabel style={font=\tiny, yshift=-2pt},
        tick label style={font=\tiny},
        ymin=0, ymax=1, xmin=0, xmax=10,
        xtick=\empty, ytick=\empty,
        axis lines=left,
        clip=false
    ]
        \addplot [dashed, gray!70, thick, domain=0:10] {0.45};
        \node[anchor=south west, font=\tiny, gray!80] at (axis cs:6, 0.45) {Full-Trace Baseline};

        \addplot [smooth, line width=1.8pt, red!70!black] coordinates { (0, 0.2) (2, 0.42) (4.5, 0.78) (7, 0.35) (10, 0.2) };
        
        \draw[dashed, red!70!black, thick] (axis cs:4.5, 0) -- (axis cs:4.5, 0.78);
        \fill[red!70!black] (axis cs:4.5, 0.78) circle (2.5pt);
        \node[below, font=\tiny, text=red!70!black] at (axis cs:7.0, 0.3) {$T^*$ (committment point)};
        \node[above, font=\tiny\bfseries, text=red!70!black] at (axis cs:4.5, 0.78) {Early Peak};
    \end{axis}

    \begin{axis}[
        name=plot2,
        at={(plot2_anchor)}, anchor=west,
        width=6.2cm, height=3cm,
        title={\textbf{Case B: Persistent Uncertainty}},
        title style={font=\small, color=green!50!black, yshift=-10pt},
        xlabel={Prefix Window Size ($T$)},
        ylabel={$\mathrm{PR\text{-}AUC}(T)$},
        xlabel style={font=\tiny, yshift=2pt},
        ylabel style={font=\tiny, yshift=-2pt},
        tick label style={font=\tiny},
        ymin=0, ymax=1, xmin=0, xmax=10,
        xtick=\empty, ytick=\empty,
        axis lines=left,
        clip=false
    ]
        \addplot [dashed, gray!70, thick, domain=0:10] {0.75};
        \node[anchor=south west, font=\tiny, gray!80] at (axis cs:0, 0.75) {Full-Trace Baseline};

        \addplot [smooth, line width=1.8pt, green!55!black] coordinates { (0, 0.15) (3, 0.4) (6, 0.62) (10, 0.72) };
    \end{axis}

    \draw[arr, red!70!black] (branch.north) |- ($(plot1.west)-(0.5,0)$) 
        node[pos=0.3, left, font=\scriptsize\bfseries, xshift=-2pt] {Case A};
        
    \draw[arr, green!55!black] (branch.south) |- ($(plot2.west)-(0.5,0)$) 
        node[pos=0.3, left, font=\scriptsize\bfseries, xshift=-2pt] {Case B};

    \end{tikzpicture}%
    }
    \caption{Our framework computes token-level uncertainty signals over prefixes of an LLM reasoning trace to diagnose how and when the model fails.}
    \label{fig:main_figure}
\end{figure*}

Detecting when language models will fail on complex reasoning tasks is an ongoing challenge with immediate ramifications for deployment reliability. Existing approaches to failure detection, such as self-consistency~\cite{wang2023selfconsistency} and uncertainty quantification~\cite{kadavath2022language, farquhar2024detecting}, treat failure as a binary prediction task. These methods can be effective at detecting when a model may fail, but they do not characterize the process through which failure emerges. We propose that this process is not singular, and treating it as such limits our understanding of how models fail and our ability to respond appropriately. 

If reasoning failures develop through different processes, then a single detection strategy cannot be optimal across all cases. Consider a model that commits to a wrong approach before its reasoning trace concludes and reproduces it consistency across completions. In this case, self-consistency will incorrectly confirm the wrong answer with high confidence and additional sampling cannot recover the failure signal. Conversely, for a model that remains genuinely uncertain throughout its reasoning, aggregating across completions would be the correct approach. These two differing situations require different detection strategies, yet existing methods apply the same approach regardless. Characterizing the process through which failures develop, instead of treating failure as a binary outcome, is a prerequisite for building detection methods that can adapt accordingly. 

Characterizing how failures manifest requires observing the model's reasoning process, not just its outcome. Recent work has made progress on this through mechanistic approaches such as probing internal activations to show incorrect answers are decodable before they are expressed ~\cite{boppana2026reasoning}, distorting reasoning steps to identify causal influence on final answers ~\cite{ye2026mechanistic} and intervening on model representations to show early commitment restricts the effectiveness of corrections ~\cite{zur2025language}. These methods reveal structure in how models fail, but they require access to model weights and internal representations. This restricts them from closed-API frontier models such as GPT-4o and Gemini, where only output tokens are accessible. Without access to the model internals, the same failure structure should be observable from external metrics such as token-level signals.

We propose a framework that characterizes reasoning failures through token-level uncertainty signals on chain-of-thought traces, requiring only log probabilities from a single completion. The framework identifies two failure modes, committed failure and persistent uncertainty, each with a distinct uncertainty trajectory across the trace. Across 23 model-dataset configurations spanning five model families and four reasoning domains, the framework's falsifiable predictions hold in 20 of 23 cases. Our framework requires moderate failure rates: extremes yield unreliable classification signals, and closed-API constraints limit available log probabilities.

This paper makes the following contributions.
(1) We propose token-level uncertainty signals predict two distinct failure modes, classifying how LLM reasoning failures manifest. We empirically validate that these failure modes are falsifiable and reproducible across diverse models and tasks.
(2) We identify the commitment point: the position in a reasoning trace at which token-level uncertainty is maximally predictive of failure, marking where the model locks onto a reasoning path.
(3) We outline practical consequences for our failure framework, showing that failure mode characterization can predict when self-consistency is effective and when single completion uncertainty features provide complementary signals.

\section{Related Work}
\paragraph{Uncertainty estimation in LLMs.}LLMs are well-calibrated on multiple-choice tasks and can estimate the probability that their own answers are correct~\cite{kadavath2022language}. Semantic entropy clusters generations by meaning rather than surface form to produce an uncertainty measure for hallucination detection, at the cost of five to ten generations per query~\cite{farquhar2024detecting}. Alignment tuning has been shown to sharpen output distributions, with the Branching Factor reducing by a factor of two to five and up to an order of magnitude at the earliest positions~\cite{yang2025logprobconcentration}. These methods treat uncertainty either at the answer level or as a static property of the output distribution. We instead study how token-level uncertainty signals evolve along a reasoning trace and show that their predictive power is non-uniform.

\paragraph{Self-consistency as a failure-detection baseline.}The dominant baseline for verifying LLM reasoning is self-consistency: sampling multiple chains of thought and taking the majority-vote answer~\cite{wang2023selfconsistency}; Semantic entropy similarly requires repeated sampling~\cite{farquhar2024detecting}. These multi-completion methods are effective when model uncertainty surfaces as inter-sample disagreement, but they are structurally blind to the committed-failure regime we identify: when a model has committed to an incorrect reasoning path early in its trace, it produces the same wrong answer consistently across completions, and self-consistency cannot distinguish these cases from genuinely correct ones. Our token-level uncertainty signals are complementary to self-consistency and operate from a single completion.

\paragraph{CoT faithfulness.}Chain-of-thought prompting~\cite{wei2022chainofthought,kojima2022large} elicits step-by-step reasoning traces and substantially improves multi-step performance. The relationship between visible CoT and the model's internal computation is contested\cite{lanham2023measuring, turpin2023language,
young2026models}. 

We analyzed CoT traces produced under standard zero-shot prompting; whether these traces faithfully reflect internal computation is orthogonal to our empirical claims, which concerns structure observable in the visible trace.

\paragraph{Trace-level structure.} Recent works have characterized failure-relevant structure in reasoning traces. Trace length itself is a confidence estimator whose relationship to accuracy is altered by reasoning post-training~\cite{devic2025trace} and, the correlation between CoT length and problem complexity is brittle, arising from approximate recall of training distribution rather than adaptive computation~\cite{palod2025performative}. At the step level, the shape of the entropy trajectory across reasoning steps has been argued to be more diagnostic than its scalar magnitude~\cite{zhao2026entropy}. We differ in two ways. First, we explicitly control for the length confound through a pre-final analysis that strips tokens after the answer marker. Second, we operate at the token level over cumulative prefix windows rather at the step level, finding that magnitude features carry more predictive signal than shape alone. 

\paragraph{Concurrent work on early commitment.} A concurrent thread has established early commitment as a recognized phenomenon in LLM reasoning through several methodological lenses. Activation probing on large reasoning models reveals that final answers are decodable from internal activations well before verbalization~\cite{boppana2026reasoning}; 
counterfactual corruption identifies a reasoning horizon at $70-85\%$ of chain length~\cite{ye2026mechanistic}; 
resampling identifies forking tokens with non-uniform importance~\cite{bigelow2025forking}; activation interventions are most effective before commitment~\cite{zur2025language}. We complement this thread along three axes. First, we operate on token-level uncertainty signals extractable from logprobs alone, without requiring model weights, counterfactual interventions, or repeated sampling. This approach makes the method deployable on closed-API models within logprob constraints, as we demonstrate on GPT-4o and Gemini-2.5Pro. Second, we characterize two qualitatively distinct failure modes, committed and persistent, with bidirectional statistical validation across $23$ (model, dataset) configurations. Third, we extend the analysis to standard inference-mode CoT models, complementing the reasoning-model focus of prior work in this thread.

\section{Methods} 
A language model's chain-of-thought reveals how it produced its final answer and, in well-calibrated cases, should be informative of whether that answer is incorrect~(Figure~\ref{fig:main_figure}). We analyze the token-level uncertainty signals across reasoning traces to characterize the structure of model failures.

\subsection{Failure Modes in LLM Reasoning}
We define model failure as any trace in which a model's final extracted answer is incorrect. If the structure of a model's reasoning determines eventual failure, then the progression of token-level signals across that trace should be characteristic of how and when failure occurs. 

We propose that this progression takes one of two qualitatively different forms. In the first failure mode, \emph{committed failure}, the model locks onto an incorrect reasoning path early in its trace. Failure becomes apparent early in the model's reasoning, and its uncertainty signals are most informative over a prefix of the trace rather than the full sequence. In the second, \emph{persistent uncertainty}, the model never commits to a reasoning path. Uncertainty builds monotonically throughout the trace, and a complete reasoning path is required to distinguish failed from successful traces. These two modes produce qualitatively different signatures in how uncertainty progresses across a reasoning trace, which we will formalize and test empirically. 

\subsection{Commitment Point}
If a model locks onto a reasoning path early, there likely is a token position where this is observable. We define this position as the commitment point: the point in a reasoning trace at which the uncertainty signals are most informative of model failure.

Beyond the commitment point, the model has already selected a reasoning path, and subsequent uncertainty is downstream noise rather than signal about the eventual outcome. In the persistent uncertainty regime, no such commitment point exists as predictive power increases monotonically, and the full trace remains more informative than any prefix. 

\subsection{Uncertainty Features}
If a model has locked onto a reasoning path, its token distribution should reflect its diminished uncertainty as the model is no longer exploring multiple paths. To reveal these failure patterns, we compute the following signals over prefixes of the reasoning trace, which we formalize below as early windows.  

Let $p^{(t)} = (p^{(t)}_1, p^{(t)}_2, \ldots)$ denote the token probability distribution at position $t$, with $p^{(t)}_{(1)} \geq p^{(t)}_{(2)} \geq \cdots$ be the sorted probabilities. For a reasoning trace of length $L$, we define the early window $\mathcal{W}_T = \{1, \ldots, \min(T, L)\}$ and compute the following uncertainty signals at each token position $t$.
\begin{description} [leftmargin=1.2cm, labelwidth=1.1cm, labelsep=0.1cm, itemsep=2pt]
    \item[\textbf{Entropy}] Spread of the top-$K$ distribution~\cite{kadavath2022language}: \\
    $\mathcal{H}_t = -\sum_i p^{(t)}_i \log p^{(t)}_i$

    \item[\textbf{Margin}] Difference between top two probabilities ~\cite{scheffer2001active}: \\
    $\mathcal{M}_t = p^{(t)}_{(1)} - p^{(t)}_{(2)}$

    \item[\textbf{NLL}] Confidence in the top token: \\
    $\mathcal{L}_t = -\log p^{(t)}_{(1)}$

    \item[\textbf{Nucleus}] Tokens needed to capture a probability threshold of $0.9$ ~\cite{holtzman2019curious}: \\
    $\mathcal{N}_t = \min \{ k : \sum_{i=1}^{k} p^{(t)}_{(i)} \geq 0.9 \}$

    \item[\textbf{Near-Tie}] Fraction of top-$K$ within $90\%$ of 
    $p^{(t)}_{(1)}$: \\
    $\mathcal{T}_t = \frac{1}{K}\sum_{i=1}^{K} \mathbf{1}[p^{(t)}_{(i)} \geq 0.9 \cdot p^{(t)}_{(1)}]$
\end{description}

For each signal $s_t \in \{\mathcal{H}_t, \mathcal{M}_t, \mathcal{L}_t, 
\mathcal{N}_t, \mathcal{T}_t\}$ and window $\mathcal{W}_T$, we compute both the mean and maximum features:
\[
\bar{s}_T = \frac{1}{|\mathcal{W}_T|}\sum_{t \in \mathcal{W}_T} s_t
\qquad \text{and} \qquad
s^{\max}_T = \max_{t \in \mathcal{W}_T} s_t
\]
Mean features capture average uncertainty across the early window and the maximum features capture the peak uncertainty event. These ten aggregated features make up the input to our classifier. 

\subsection{Diagnosing Failure Modes with PR-AUC}
To quantify failure mode detection, we compute PR-AUC, a measure of a classifier's ability to identify failures, at each early window and examine the shape of the resulting curve. We propose that each failure mode will differ along the following axes: the shape of the PR-AUC curve across window sizes, whether a commitment point $T^*$ exists and whether the $95\%$ bootstrap confidence interval shows the early window is more informative than the full trace. 

The committed failure mode should reveal an inverted-U shape in its PR-AUC plot. As the early window expands, the predictive power of the uncertainty signals should grow as well. The PR-AUC reaches its peak at the commitment point, $T^*$ and subsequent early windows decline in their predictive performance, producing an inverted-U curve. We further categorize committed failures into \emph{strong committed} and \emph{weak committed} based on the bootstrap confidence interval on $\Delta(T^*) = \mathrm{PR\text{-}AUC}(T^*) -\mathrm{PR\text{-}AUC}(\mathrm{full})$, the difference between the peak early-window PR-AUC and the full-trace PR-AUC. Strong committed failures have a $95\%$ confidence interval on $\Delta(T^*)$ that excludes zero, confirming that the early window is strictly more informative than the full trace. Weak committed failures have $\Delta(T^*) > 0$. The early windows still outperform the full trace in expectation, but confidence intervals span zero due to statistical uncertainty.  

In contrast, the PR-AUC of persistent uncertainty regimes monotonically rises. The PR-AUC curve steadily increases and never exceeds that of the full trace because the model never selects a path and each subsequent token continues to add genuine signal. Formally, $\Delta(T) < 0$ for all tested windows $T$, revealing that no early window can recover the full information available in the complete trace. This shows the absence of a commitment event; there is no position along a reasoning trace that has concentrated commitment power.  

\subsection{Pre-final Analysis}
The length of a model's reasoning trace often correlates with the correctness of its final answer \citep{devic2025trace}. Failing models tend to write longer traces, inflating full-trace uncertainty features by proxying trace length rather than capturing genuine uncertainty. To account for this potential length confound, we strip all tokens that occur after the final answer marker in the reasoning trace before computing the uncertainty signals, an approach we denote as pre-final analysis. In configurations where failing models produce substantially longer traces, a length confound could inflate the full-trace PR-AUC curve, omitting an otherwise underlying inverted-U signature. The pre-final analysis is designed to prevent this by controlling for post-answer token length within each configuration. 

\subsection{Connection to Self-Consistency}
The framework's failure-mode classification has direct implications for self-consistency~\citep{wang2023selfconsistency}, which samples $k$ completions and uses majority-vote agreement as a confidence signal. In the committed regime, a model can reproduce the same wrong answer consistently across completions, making agreement rate an unreliable signal; single-completion uncertainty captures within-completion structure that agreement cannot observe. In the persistent regime, failure signatures surface as cross-completion disagreement, and self-consistency aggregation is genuinely informative. Section~\ref{sec:results} empirically tests both regime-conditional triage and complementarity.

\section{Experiments}
We test whether the two failure modes manifest empirically across a range of models, datasets and task difficulty levels. Our framework operates entirely on the externalized chain-of-thought trace and requires only token-level log probabilities; no access to internal model representations is needed. Our code is publicly available.\footnote{\url{https://github.com/sisl/LMTwoFailureModeFramework}}

\subsection{Models and Datasets}
We evaluate models spanning a range of sizes, families and architectures: Qwen3.5-2B, Qwen3.5-9B, Qwen3.5-27B, Qwen3.5-122B-A10B~\cite{team2026qwen3}, Llama3.1-8B-Instruct ~\cite{grattafiori2024llama}, GPT-OSS-20B ~\cite{agarwal2025gptoss}, Gemma4-31B, GPT-4o ~\cite{achiam2023gpt}, Gemini-2.5Pro ~\cite{comanici2025gemini}. We include both dense models, mixture of experts, open-source and frontier models in order to capture a variety of patterns. These are evaluated on five benchmarks spanning mathematics, scientific, logical and coding domains: GSM8K ($1319$ test questions of grade-school math; \citealt{cobbe2021trainingverifierssolvemath}), MATH-500 ($500$ competition-level math problems representative of the full MATH benchmark; \citealt{hendrycks2021math}, \citealt{lightman2024lets}), GPQA Diamond ($198$ multiple-choice questions on graduate-level biology, chemistry and physics; \citealt{rein2024gpqa}), and LiveCodeBench ($451$ applicable coding challenges; \citealt{jain2025livecodebench}). We additionally evaluated AR-LSAT ($230$ questions from the Law School Admissions Test; \citealt{zhong2021arlsat}) but every configuration we ran fell outside the applicability band; these results are reported in Table~\ref{tab:excluded} in the appendix,  for transparency and excluded from the pool as a scope decision. These datasets were selected to span across domains and a range of difficulty levels relative to model capability, with the intention to generalize the failure framework. Failure rates below $15\%$ or above $60\%$ paired with an AUROC $<0.55$ render the framework inapplicable. We additionally exclude configurations whose prefinal-stripped trace contains too few failures to support reliable analysis (typically fewer than $\sim$10 failures in the prefinal-valid subset), where the analysis pipeline falls back to regular features. 

All experiments use a temperature of $0.6$, which is consistent with standard LLM evaluation practice \citep{renze2024temperature} and balances exploration and exploitation. For open-weight models, we retrieve the top $200$ log probabilities per token (capture around $99\%$ of all probability mass) since almost all probability mass is concentrated on a small subset of tokens \citep{yang2025logprobconcentration}. The frontier models, GPT-4o and Gemini2.5-Pro expose only the top $20$ log probabilities, so their experiments are correspondingly constrained. We do not test Claude models as they do not expose any log probabilities at the time of testing. All open-source models are served using vLLM ~\cite{kwon2023efficient}. 

Models are prompted to format their final answer within \texttt{\string\boxed\{\}} for math datasets or \textit{Final: Answer} for GPQA Diamond, and answers are extracted via regex. All experiments were run on 2 × NVIDIA H100 96GB GPUs, for a total of approximately 200 GPU hours across all configurations. 

\subsection{Evaluation Protocol}
For each question, we prompt the model with the instruction: \textit{``Reason through the problem step by step to arrive at an answer''} \citep{wei2022chainofthought}, and compute uncertainty signals over the resulting reasoning trace. We define a binary failure label $y = \neg \text{correct}$, where the correctness is determined by the automated extraction of the final answer.

We compute the uncertainty signals over prefixes of each trace, where a prefix is the first $T$ tokens in the trace with $T \in \{128, 256, 400, 512, 1024, 2048\}$. A single inference call is made per question and the features for each window are computed over the same trace, ensuring that comparisons across $T$ are not confounded by sampling variation. 

We use PR-AUC as our primary evaluation metric. Across our experiments, model failure rates range from $5\%$ to $84\%$. At these class priors, AUROC can remain high even when a classifier has poor precision on the minority class \citep{davis2006roc}, making PR-AUC more informative for evaluating failure detection. 

To calculate PR-AUC, we use out-of-fold (OOF) predictions from a 5-fold stratified logistic regression classifier with balanced class weights and regularization strength $C=1.0$. The OOF predictions are concatenated across all folds to obtain a single set of predictions: $\{(\hat{p}_i, y_i)\}_{i=1}^{n}$, where $y_i \in \{0, 1\}$ indicates failure. The random baseline PR-AUC is the empirical failure rate 
$\bar{y} = \frac{1}{n}\sum_{i=1}^{n} y_i$. Statistical uncertainty is quantified via paired bootstrap confidence intervals on 
\begin{equation}
    \Delta\text{PR-AUC} = \text{PR-AUC}_{\text{early}} - 
    \text{PR-AUC}_{\text{full}}
    \label{eq:delta_prauc}
\end{equation}
resampling $n$ observations with replacement for $10{,}000$ iterations. 

To test whether the inverted-U signature truly reproduces across model-dataset configurations, we pool per-configuration evidence using four complementary tests: a sign test on the direction of $\hat{\Delta}$ across committed configurations, Stouffer's $Z$ combining bootstrap $p$-values, an inverse-variance weighted meta-analysis estimating a pooled effect size and a joint sign test evaluating the framework's bidirectional prediction across all configurations simultaneously. Classification into committed or persistent is made for each configuration independently before pooling. In cases where within-dataset stratification reveals that an aggregate $\Delta$PR-AUC averages over distinct modes, difficulty strata are substituted as units.

The four tests are complementary as each answers a distinct question about either directional consistency, cumulative significance, pooled magnitude and bidirectional falsifiability. 

Several configurations are excluded from our primary analysis based on two methodological criteria: (i) the failure rate falls outside $[15\%, 60\%]$ and the AUROC falls below $0.55$, or (ii) the prefinal-stripped trace contains too few failures for reliable analysis. AR-LSAT is additionally excluded as a scope decision.

\subsection{Commitment Point Identification}
A commitment point is identified as the window achieving the highest PR-AUC whose lower bound of the 95\% bootstrap confidence interval on $\Delta\text{PR-AUC}$ excludes zero (\emph{strong committed}), or when $\Delta\text{PR-AUC} > 0$ with $p(\Delta > 0) > 0.8$ but the interval spans zero (\emph{weak committed}).

We restrict identification of the commitment point to genuine early windows. Windows that capture the majority of the trace are excluded, as they no longer constitute an early observation of the reasoning process. Concretely, this results in early windows of $1024$ and $2048$ being omitted if the maximum length of a reasoning trace is less than either threshold. 

\subsection{Evaluation Against Self-Consistency}
We compare how single-completion uncertainty signals relate to self-consistency. Specifically, we identify when self-consistency is effective and whether uncertainty signals add predictive power on top of when it is. We evaluate three models over $k=15$ completions each on GPQA Diamond, Gemma4-31B (weak commitment), Qwen3.5-9B (persistent uncertainty) and Qwen3.5-122B (persistent uncertainty), spanning the two failure regimes, model families and model sizes.  

\section{Results}\label{sec:results}
Across $23$ model and dataset configurations, we find clear evidence for both failure modes in our framework. Fourteen configurations exhibit committed failure, where the early uncertainty signals reach peak predictive power before the full trace. Nine exhibit persistent uncertainty, where the PR-AUC accumulates monotonically and never surpasses that of the full trace. Additionally, we verify that reasoning trace lengths do not differ systematically across the two failure regimes, ruling out trace length as a confound on the failure mode framework itself. 




\begin{table}[htbp]
  \centering
  \resizebox{\columnwidth}{!}{%
  \begin{tabular}{@{}
    l
    S[table-format=2.2(2)]
    S[table-format=2.2(2)]
    S[table-format=2.2(2)]
    S[table-format=2.2(2)]
    S[table-format=2.2(2)]
    @{}
  }
    \toprule
     & \multicolumn{4}{c}{Benchmark} \\
    \cmidrule(l){2-5}
    {Model}                  & {GSM8K} & {MATH500} & {GPQA} & {LiveCodeBench} \\
    \midrule
   Qwen3.5-2B               & {WC}      & {WC}         & {WC}      & {Persist}       \\
  Qwen3.5-9B               & {-}       & {-}          & {Persist} & {-}             \\
  Qwen3.5-122B-A10B        & {-}       & {WC}         & {Persist} & {Persist}       \\
  Llama3.1-8B-Instruct     & {WC}      & {Persist}    & {-}       & {-}             \\
  GPTOSS-20B               & {Persist} & {WC}         & {Persist} & {WC}            \\
  Gemma4-31B               & {-}       & {Stratified} & {WC}      & {SC}            \\
  GPT-4o                   & {-}       & {Persist}    & {WC}      & {-}             \\
  Gemini-2.5Pro            & {-}       & {SC}         & {-}       & {-}             \\
    \bottomrule
  \end{tabular}
  }
\caption{Failure-mode classification by (model, dataset) configuration. \emph{SC}/\emph{WC}: strong/weak committed (95\% CI on $\Delta(T^*)$ excludes/spans zero). \emph{Persist}: persistent uncertainty. \emph{Stratified}: per-difficulty-level analysis. $-$: excluded (see Table~\ref{tab:excluded}).}
  \label{tab:failure_types}
\end{table}


\vspace{-8pt}
\subsection{Committed Failures}
Committed failure is the most prevalent failure mode in our experiments, occurring in fourteen model-dataset configurations across five model families, four datasets and a range of model scales. All committed cases have an inverted U-shape PR-AUC curve where the predictive power increases as the early window grows to include the commitment point $T^*$, after which it falls as additional tokens dilute the early signal. 

\begin{figure}[t]
    \centering
    \includegraphics[width=\columnwidth]{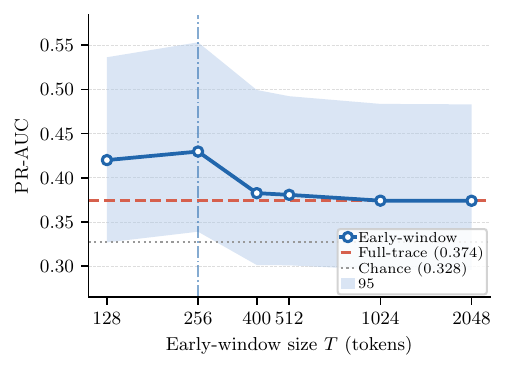}
    \caption{Strong Committed Failure: Gemma4-31 on LiveCodeBench. The $\Delta(T)$ confidence interval excludes $0$.}
    \label{fig:commit_gemma4_lcb}
\end{figure}

The strongest committed failure signatures occur when $\Delta(T^*)$ excludes zero, confirming that the early window trace is strictly more informative than the full trace. This strong committed signature is most evident for Gemma4-31b on LiveCodeBench, where the inverted-U shape is visually unambiguous and the CI excludes zero with high confidence~(Figure~\ref{fig:commit_gemma4_lcb}). GPT-OSS-20B on the easy-question split of LiveCodeBench independently replicates this signature with the cleanest delta confidence bands across four consecutive committed windows. Gemini-2.5-Pro on MATH-500 further confirms this pattern, though this result uses the top-$20$ log probabilities rather than the full output distributions. These cases demonstrate the committed failure regime clearly: the model selects an incorrect reasoning path before the trace finishes, and subsequent tokens add noise relative to the early signal. 
Not all committed failures appear as strongly. Several configurations demonstrate the inverted U-shape and a positive $\Delta(T)$ but with confidence intervals that span zero. We classify these failures as \emph{weak committed}. This continuum of signal strength arises when the failure pool is small (Gemma4-31b on GPQA with $43$ failures), when the curve collapses sharply after the commitment point (Qwen3.5-2B on GPQA, where $\Delta$ peaks at $T^*{=}512$ before dropping below the full-trace baseline at $T{=}1024$) or when a high failure rate compresses the signature of the uncertainty features (GPT-OSS-20B on LCB hard at a $60\%$ failure rate). The commitment point still exists across all these cases, and the statistical uncertainty is a reflection of sample constraints. 

These examples generalize across all three tested axes of variation (Table~\ref{tab:failure_types}): four reasoning domains (GSM8K, MATH-500, GPQA, LiveCodeBench), five model families (Gemini, Gemma, GPT-OSS, Llama, Qwen), and a model-scale range from Qwen3.5-2B through frontier-scale systems. No single architecture, training pipeline, or task family explains the inverted-U pattern.

\subsection{Persistent Uncertainty Failures}
Persistent uncertainty appears in nine of the model-dataset configurations, revealing a diagnostically distinct pattern from committed failure. Instead of an inverted-U, the PR-AUC curve rises monotonically with window size, and the full trace is consistently higher than any early window. There is no $T^*$ at which subsequent tokens become noise since the full trace outperforms all windows. Additional tokens always add signal because the model does not have a single position with concentrated predictive power throughout the trace. The model is genuinely uncertain throughout its reasoning, and the uncertainty is only fully understood once we observe the full trace. 

\begin{figure}[t]
    \centering
    \includegraphics[width=\columnwidth]{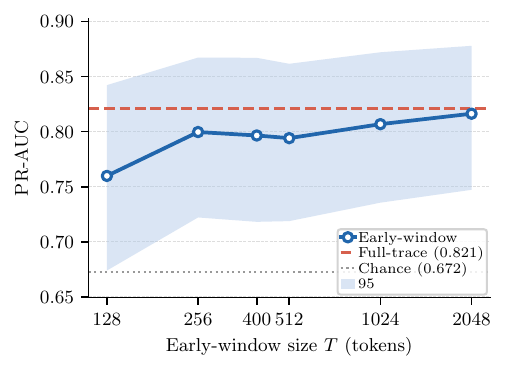}
    \caption{Persistent Uncertainty: Llama3.1-8B on MATH-500. Early windows never beat the full trace.}
    \label{fig:persist_llamaMath500}
\end{figure}

The clearest example of this is Llama3.1-8b on MATH-500 where the PR-AUC rises steadily across all window sizes with no peak~(Figure~\ref{fig:persist_llamaMath500}). The model does not lock onto a wrong path but instead searches paths across the full trace, with uncertainty remaining elevated throughout. 
This is not a phenomenon restricted to model architecture or task type. It appears in GPT-OSS-20B on GSM8K, a dataset which other models exhibited committed failure on. We also observe it in Qwen3.5-122B on GPQA and LiveCodeBench, the largest model in our evaluation. In all persistent uncertainty cases the diagnostic is the same. 

GPT-4o presents a nuanced case under the API's top-$20$ log probability constraint. On MATH-500, the pattern is persistent uncertainty: $\Delta$ is uniformly negative across all early windows and the full trace dominates. On GPQA, the pattern is weak committed ($\hat{\Delta} = +0.005$ at $T^*=1024$); near the persistent boundary. The top-$20$ constraint affects feature reliability: mean-based features (entropy, NLL) depend on tail probability mass that is missing, while max-based features (margin, near-tie) remain valid since they depend only on the single-highest probability token. Despite this constraint, both configurations are in the pool with classifications consistent with the framework's prediction.

\subsection{Reproducibility Across Configurations}
\begin{table}[h!]
    \centering
    \small
    \setlength{\tabcolsep}{6pt}
    \begin{tabular}{@{}lr@{}}
    \toprule
    Test & Result \\
    \midrule
    Sign test (committed)       & 14/14, $p = 6.1\times10^{-5}$ \\
    Joint sign test (all)       & 20/23, $p = 2.4\times10^{-4}$ \\
    Stouffer's $Z$ (committed)  & $Z=5.48$, $p = 2.1\times10^{-8}$ \\
    IV-weighted $\hat{\Delta}$  & $+0.013$ $[{+0.005}, {+0.020}]$ \\
    \bottomrule
    \end{tabular}
\caption{Meta-analytic pooling of $\Delta(T^*)$ across configurations. Stouffer's $Z$ combines per-configuration evidence on the standard-normal scale; the joint sign test pools the committed and persistent classes.}
    \label{tab:pooling_headline}
\end{table}
The framework's directional predictions are entirely consistent in the committed regime: every configuration shows $\hat{\Delta} > 0$ as predicted (Table~\ref{tab:pooling_headline}, Figure~\ref{fig:forest}). Persistent configurations follow the opposite prediction in six of nine cases; the three boundary cases (Gemma4-31B / MATH-500 L5, Qwen3.5-2B / LCB, Qwen3.5-122B / LCB) have $\hat{\Delta}$ within $\pm 0.003$ of zero --- statistically indistinguishable from the prediction rather than violations. The framework is falsifiable across both signs and survives.

\subsection{Self-Consistency}
We find that triage performance tracks failure mode classification directly~(Figure~\ref{fig:triagecomparison}; full operating curves in Appendix Table~\ref{tab:sc_triage}). Pre-final stripping is essential across all models since the full trace signal collapses as a triage tool in every panel. With pre-final features, Gemma4-31B (committed) achieves near-perfect recall up to a $30\%$ skip rate using either the prefinal early window ($T=400$) or the full prefinal trace. For both Qwen models in the persistent regime, the pre-final curves hold near $1.0$ in the top $20\%$ of confident questions skipped and degrade more steeply beyond. For Qwen3.5-9B, the early-window and full-prefinal curves are indistinguishable, reaffirming the absence of a commitment point.

\begin{figure}[h]
    \centering
    \includegraphics[width=\columnwidth]{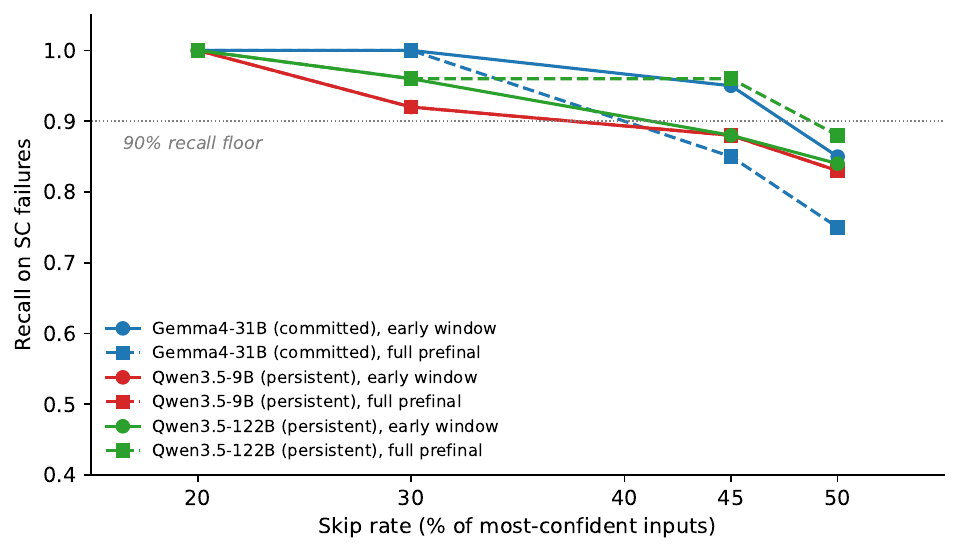}
\caption{Selective self-consistency triage: recall on SC failures vs. skip rate (\% of most-confident inputs skipped), for committed (Gemma4-31B / GPQA) and persistent (Qwen3.5-9B / GPQA) configurations. Pre-final stripping is essential --- the full-trace curve collapses without it.}
    \label{fig:triagecomparison}
\end{figure}

We then evaluate whether uncertainty signals complement self-consistency's agreement rate. Single-completion uncertainty alone is substantially weaker than agreement rate across all three configurations (PR-AUC $\approx 0.42$ vs.\ $\approx 0.78$), confirming that the two signals are not interchangeable. Combining them improves PR-AUC over agreement alone in every configuration: $+0.026$ (Gemma4-31B), $+0.035$ (Qwen3.5-9B), $+0.045$ (Qwen3.5-122B). While individual-cell lifts have CIs that span zero given GPQA's small failure pool ($n=198$), the consistent positive direction across both regimes indicates that uncertainty features capture within-completion reasoning quality that agreement rate, a purely between-completion signal, cannot observe (Appendix Figure~\ref{fig:sc_complementarity}). When self-consistency is already deployed, adding uncertainty features incurs no additional inference cost and consistently improves failure prediction in expectation.

\section{Conclusion}
We introduced a two-mode framework characterizing how language-model reasoning failures manifest in chain-of-thought traces, requiring only log probabilities from a single completion. Across 23 configurations spanning five model families and four reasoning domains, the framework's bidirectional prediction holds in 20 of 23 cases (sign test on committed configurations: 14/14, $p = 6.1{\times}10^{-5}$; pooled $\hat{\Delta} = +0.013$, 95\% CI $[+0.005, +0.020]$). 

The failure-mode classification has direct deployment implications: in the committed regime we can skip self-consistency on the top 30\% most-confident inputs without sacrificing failure recall, and across both regimes, combining uncertainty features with the agreement rate yields a consistent positive lift. Failure detection strategies should be adapted to failure mode rather than applied uniformly.

\section{Limitations}
Our framework requires failure rates within a workable applicability band; configurations at the extremes do not produce reliable PR-AUC estimates and are excluded from the pool. We use a single completion per question, so commitment-point identification is sensitive to the sampled trace. The commitment point $T^*$ is identified at the granularity of six fixed window sizes ($\{128, 256, 400, 512, 1024, 2048\}$) and represents a window range rather than an exact token position; a finer-grained sweep between adjacent windows could refine its localization. Closed-API constraints, where GPT-4o and Gemini-2.5Pro expose only their top-20 log probabilities, limit the reliability of mean-based features that depend on tail probability mass; max-based features remain valid under truncation. Our self-consistency evaluation is restricted to three configurations on a single benchmark (GPQA Diamond), so consistency of the complementarity result across broader settings remains to be validated. Finally, the framework operates on visible chain-of-thought traces and does not address whether these traces faithfully reflect internal model computation; this question is orthogonal to our empirical claims, which concern structure observable in the visible trace.

\bibliography{custom}
\appendix
\section*{Appendix}
\label{sec:appendix}

Additional results and excluded configurations are reported below.

\begin{table*}[!h]
\centering
\small
\setlength{\tabcolsep}{5pt}
\begin{tabular}{@{}lr@{}}
\toprule
{Configuration} & {Exclusion reason} \\
\midrule
Qwen3.5-9B / GSM8K         & Capability ceiling ($1.1\%$ failure rate) \\
Qwen3.5-9B / MATH-500      & Below band ($14.6\%$ failure, AUROC $0.43$) \\
Llama-3.1-8B / GPQA        & Capability floor ($63.4\%$ failure, AUROC $0.40$) \\
Gemini-2.5Pro / GPQA       & Prefinal-valid subset too small ($18/198$, 4 failures) \\
Llama-3.1-8B / LCB         & Prefinal-valid subset empty ($0/455$) \\
Qwen3.5-9B / LCB           & Prefinal-trace confound (no pipeline output) \\
AR-LSAT / all models (4)   & Scope: legal/analytical reasoning \\
\bottomrule
\end{tabular}
\captionof{table}{The framework's applicability band is $[15\%, 60\%]$ failure rate paired with AUROC $> 0.55$; below $15\%$ the PR-AUC estimator is unstable due to insufficient positive class examples, above $60\%$ capability-floor noise dominates the signal. The prefinal-trace confound criterion excludes configurations whose prefinal-valid subset contains fewer than $\sim$10 failures, where the analysis pipeline falls back to regular features and the prefinal-mode comparison becomes unreliable. AR-LSAT is excluded as a scope decision rather than treated as evidence against the framework.}
\label{tab:excluded}
\end{table*}

\begin{table*}[h!]
\centering
\small
\setlength{\tabcolsep}{6pt}
\begin{tabular}{@{}lrrrr@{}}
\toprule
{Scenario} & $N$ & {IV-weighted} ${\hat{\Delta}}$ & {Stouffer's} $p$ & {Joint sign test} \\
\midrule
A. Gemma stratified \emph{(primary)}     & 23 & $+0.013$ {\footnotesize [+0.005, +0.020]} & $2.1{\times}10^{-8}$ & 20/23, $p = 2.4{\times}10^{-4}$ \\
B. Gemma aggregate (no stratification)   & 20 & $+0.012$ {\footnotesize [+0.004, +0.019]} & $2.0{\times}10^{-6}$ & 17/20, $p = 1.3{\times}10^{-3}$ \\
C. Include prefinal-confound cells       & 25 & $+0.013$ {\footnotesize [+0.005, +0.020]} & $2.1{\times}10^{-8}$ & 22/25, $p = 1.1{\times}10^{-4}$ \\
\bottomrule
\end{tabular}
\caption{Robustness of the pooled estimate to two analytic choices: how Gemma4-31B / MATH-500 is entered (Scenario~A: stratified by difficulty level, the primary choice motivated by between-mode cancellation in the aggregate; Scenario~B: aggregate), and whether configurations failing the prefinal-trace confound criterion are excluded (A, B) or included (Scenario~C). The pooled effect size and statistical evidence are stable across all three choices.}
\label{tab:sensitivity}
\end{table*}

\begin{figure*}[t]
    \centering
    \includegraphics[width=\textwidth]{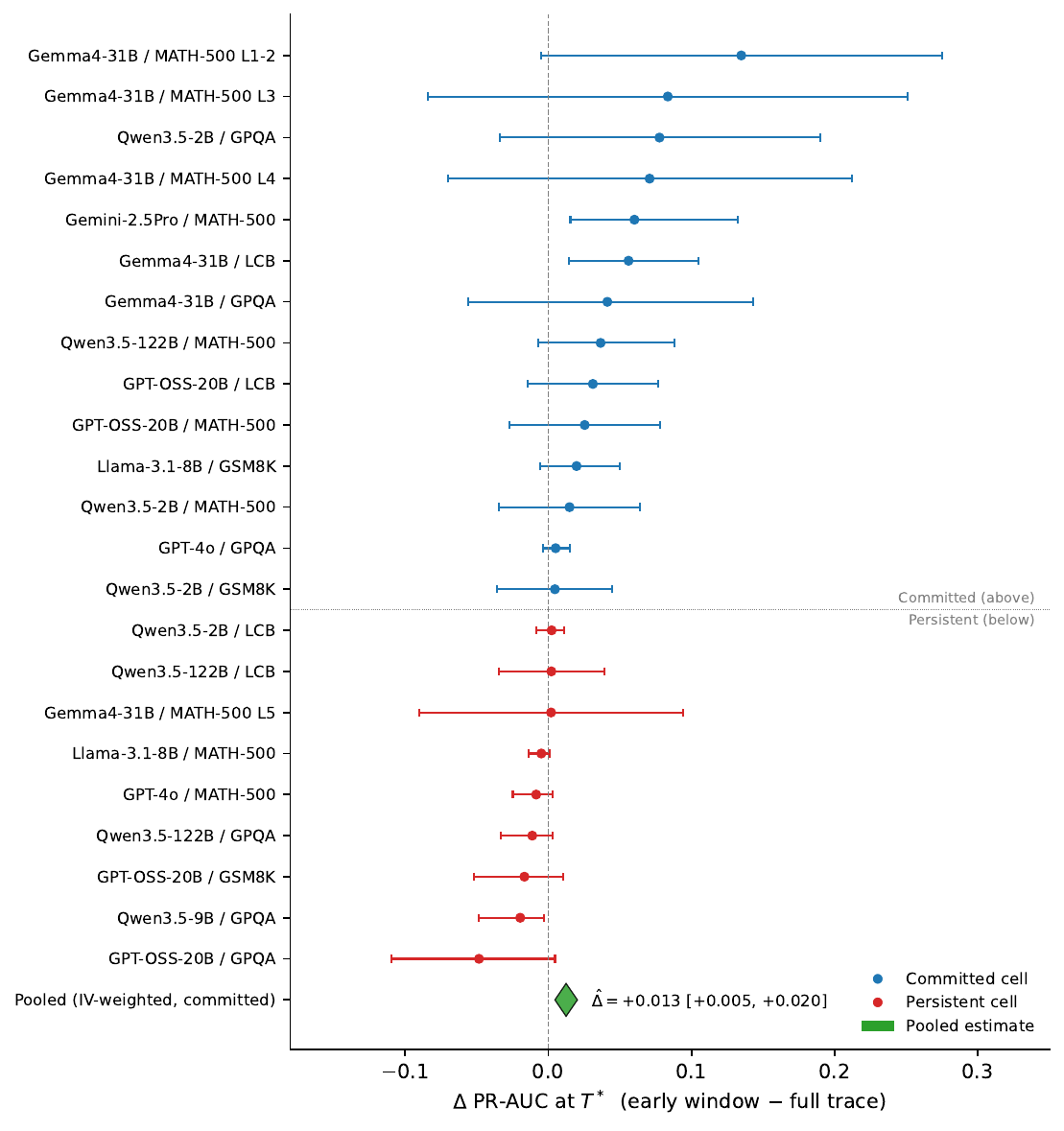}
    \caption{Forest plot of $\Delta$PR-AUC at $T^*$ across all 23 configurations with $95\%$ bootstrap CIs. Committed configurations (blue, upper group) all show positive $\Delta$, ranging from $+0.005$ to $+0.135$. Persistent configurations (red, lower group) cluster near or below zero, with three boundary cases (Gemma4-31B / MATH-500 L5, Qwen3.5-2B / LCB, Qwen3.5-122B / LCB) at $\hat{\Delta} \approx +0.002$. The green diamond is the inverse-variance weighted pooled estimate over committed configurations ($\hat{\Delta}=+0.013$, $95\%$ CI $[+0.005, +0.020]$).}
    \label{fig:forest}
\end{figure*}

\begin{table*}[t]
\centering
\small
\setlength{\tabcolsep}{5pt}
\begin{tabular}{@{}lrrrrrr@{}}
\toprule
{Signal} & {Skip 20\%} & {Skip 30\%} & {Skip 45\%} & {Skip 50\%} & {Skip for $\ge 90\%$ recall} \\
\midrule
\multicolumn{6}{l}{\textit{Gemma4-31B / GPQA (committed; SC failures: 20/196, 10.2\%)}} \\
\quad Full trace (no prefinal strip) & $0.65/0.82$ & $0.65/0.88$ & $0.55/0.90$ & $0.45/0.89$ & $10\%$ skip \\
\quad Prefinal full trace            & $1.00/1.00$ & $1.00/1.00$ & $0.85/0.97$ & $0.75/0.95$ & $40\%$ skip \\
\quad Prefinal early ($T=400$)       & $1.00/1.00$ & $1.00/1.00$ & $0.95/0.99$ & $0.85/0.97$ & $45\%$ skip \\
\midrule
\multicolumn{6}{l}{\textit{Qwen3.5-9B / GPQA (persistent; SC failures: 24/193, 12.4\%)}} \\
\quad Full trace (no prefinal strip) & $0.75/0.84$ & $0.71/0.88$ & $0.63/0.90$ & $0.54/0.89$ & $10\%$ skip \\
\quad Prefinal full trace            & $1.00/1.00$ & $0.92/0.97$ & $0.88/0.97$ & $0.83/0.96$ & $35\%$ skip \\
\quad Prefinal early ($T=2048$)      & $1.00/1.00$ & $0.92/0.97$ & $0.88/0.97$ & $0.83/0.96$ & $35\%$ skip \\
\bottomrule
\end{tabular}
\caption{Full triage operating curves. Each table entry reports \emph{recall / precision} on the self-consistency verdict for inputs flagged by ranking single-completion confidence. Prefinal stripping (removing post-answer tokens from the trace before extracting uncertainty features) eliminates a length confound; without it, generated-length leakage produces overconfident triage at the cost of recall. Gemma4-31B (committed) sustains perfect triage out to a 30\% skip rate; Qwen3.5-9B (persistent) sustains it only to 20\%.}
\label{tab:sc_operating}
\end{table*}

\begin{table*}[h]
\centering
\small
\setlength{\tabcolsep}{4pt}
\begin{tabular}{@{}lccr@{}}
\toprule
{Operating point} & {Compute} & {Recall} & {Precision} \\
                         & {saved}   & {on SC fails} & \\
\midrule
\multicolumn{4}{l}{\textit{Committed: Gemma4-31B / GPQA}} \\
\quad Skip top 20\% & $20\%$ & $1.00$ & $1.00$ \\
\quad Skip top 30\% & $30\%$ & $1.00$ & $1.00$ \\
\quad Skip top 45\% & $45\%$ & $0.95$ & $0.99$ \\
\quad Skip top 50\% & $50\%$ & $0.85$ & $0.97$ \\
\midrule
\multicolumn{4}{l}{\textit{Persistent: Qwen3.5-9B / GPQA}} \\
\quad Skip top 20\% & $20\%$ & $1.00$ & $1.00$ \\
\quad Skip top 30\% & $30\%$ & $0.92$ & $0.97$ \\
\quad Skip top 45\% & $45\%$ & $0.88$ & $0.97$ \\
\quad Skip top 50\% & $50\%$ & $0.83$ & $0.96$ \\
\bottomrule
\end{tabular}
\caption{Self-consistency triage on GPQA. ``Skip top $k\%$'' ranks inputs by single-completion confidence and skips 15-completion SC on the $k\%$ most confident. Recall and precision are against the SC verdict. Full operating curves in Appendix~\ref{tab:sc_operating}.}
\label{tab:sc_triage}
\end{table*}

\begin{figure*}[!h]
    \centering
    \includegraphics[width=0.6\textwidth]{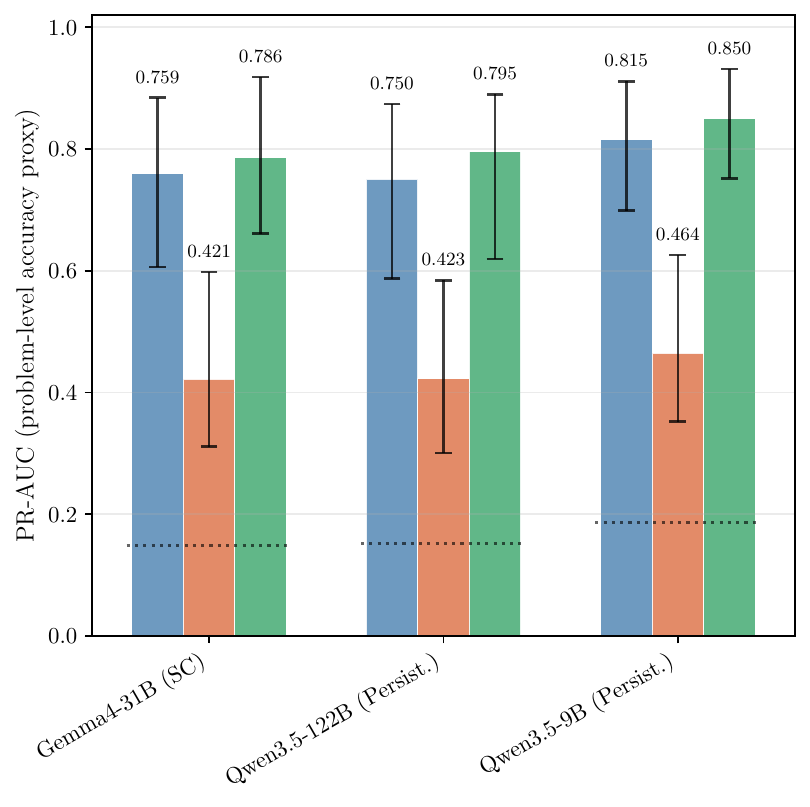}\\
    \includegraphics[width=0.8\columnwidth]{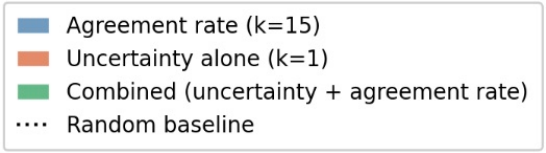}
    \caption{Comparison of self-consistency's agreement rate signal, single-completion uncertainty signals and the aggregate of the two across Gemma4-31B, Qwen3.5-9B and Qwen3.5-122B on GPQA Diamond. }
    \label{fig:sc_complementarity}
\end{figure*}





\begin{table*}[htbp]
  \centering
  \resizebox{\textwidth}{!}{%
  \begin{tabular}{@{}
    ll
    S[table-format=2.2(2)]
    S[table-format=2.2(2)]
    S[table-format=2.2(2)]
    S[table-format=2.2(2)]
    S[table-format=2.2(2)]
    S[table-format=2.2(2)]
    S[table-format=2.2(2)]
    @{}
  }
    \toprule
    \multirow{2}{*}{Benchmark} & \multirow{2}{*}{Model} & \multicolumn{7}{c}{PR-AUC}  \\
    \cmidrule(l){3-9}
                              &                      & {$T=128$}             & {$T=256$}       & {$T=400$} & {$T=512$} & {$T=1024$} & {$T=2048$}  & {Full Trace}   \\
    \midrule
    \multirow{3}{*}{GSM8K}   & Qwen3.5-2B & \shortstack{0.3859 \\ \footnotesize{[0.3324, 0.4514]}} & \shortstack{\bfseries 0.3991 \\ \footnotesize{[0.3439, 0.4637]}} & \shortstack{0.3974 \\ \footnotesize{[0.3450, 0.4608]}} & \shortstack{0.3934 \\ \footnotesize{[0.3420, 0.4563]}} & \shortstack{0.3934 \\ \footnotesize{[0.3416, 0.4579]}} & \shortstack{0.3957 \\ \footnotesize{[0.3431, 0.4606]}} & \shortstack{0.3934 \\ \footnotesize{[0.3411, 0.4586]}} \\
                              & Llama3.1-8B-Instruct     & \shortstack{0.3367 \\ \footnotesize{[0.2762, 0.4161]}} & \shortstack{0.4618 \\ \footnotesize{[0.3885, 0.5468]}} & \shortstack{0.5029 \\ \footnotesize{[0.4267, 0.5848]}} & \shortstack{\bfseries 0.5067 \\ \footnotesize{[0.4309, 0.5872]}} & \shortstack{0.4818 \\ \footnotesize{[0.4064, 0.5679]}} & \shortstack{0.4829 \\ \footnotesize{[0.4071, 0.5690]}} & \shortstack{0.4841 \\ \footnotesize{[0.4079, 0.5700]}} \\
                              & GPTOSS-20B               & \shortstack{0.0963 \\ \footnotesize{[0.0578, 0.1765]}} & \shortstack{0.1420 \\ \footnotesize{[0.0902, 0.2411]}} & \shortstack{0.1471 \\ \footnotesize{[0.0937, 0.2450]}} & \shortstack{0.1398 \\ \footnotesize{[0.0905, 0.2308]}} & \shortstack{0.1558 \\ \footnotesize{[0.1031, 0.2502]}} & \shortstack{\bfseries 0.1760 \\ \footnotesize{[0.1166, 0.2784]}} & \shortstack{0.1914 \\ \footnotesize{[0.1252, 0.3018]}} \\

    \midrule
    \multirow{17}{*}{MATH500} & Qwen3.5-2B               & \shortstack{0.2665 \\ \footnotesize{[0.2140, 0.3479]}} & \shortstack{0.2602 \\ \footnotesize{[0.2095, 0.3369]}} & \shortstack{0.3048 \\ \footnotesize{[0.2398, 0.3872]}} & \shortstack{0.3254 \\ \footnotesize{[0.2567, 0.4207]}} & \shortstack{\bfseries 0.3535 \\ \footnotesize{[0.2778, 0.4545]}} & \shortstack{0.3152 \\ \footnotesize{[0.2487, 0.4079]}} & \shortstack{0.3388 \\ \footnotesize{[0.2653, 0.4379]}} \\
                              & Qwen3.5-9B & \shortstack{0.1109 \\ \footnotesize{[0.0762, 0.1691]}} & \shortstack{0.1535 \\ \footnotesize{[0.1015, 0.2371]}} & \shortstack{\bfseries 0.1591 \\ \footnotesize{[0.1051, 0.2558]}} & \shortstack{0.1591 \\ \footnotesize{[0.1038, 0.2537]}} & \shortstack{0.1302 \\ \footnotesize{[0.0873, 0.1997]}} & \shortstack{0.1304 \\ \footnotesize{[0.0874, 0.1998]}} & \shortstack{0.1304 \\ \footnotesize{[0.0874, 0.1998]}} \\
                              & Qwen3.5-122B-A10B        & \shortstack{0.1441 \\ \footnotesize{[0.1134, 0.1871]}} & \shortstack{0.1850 \\ \footnotesize{[0.1405, 0.2559]}} & \shortstack{\bfseries 0.2109 \\ \footnotesize{[0.1552, 0.3005]}} & \shortstack{0.1969 \\ \footnotesize{[0.1470, 0.2746]}} & \shortstack{0.1800 \\ \footnotesize{[0.1361, 0.2515]}} & \shortstack{0.1755 \\ \footnotesize{[0.1342, 0.2446]}} & \shortstack{0.1755 \\ \footnotesize{[0.1342, 0.2446]}} \\
                              & Llama3.1-8B-Instruct     & \shortstack{0.7598 \\ \footnotesize{[0.6739, 0.8422]}} & \shortstack{0.7996 \\ \footnotesize{[0.7221, 0.8672]}} & \shortstack{0.7965 \\ \footnotesize{[0.7181, 0.8670]}} & \shortstack{0.7941 \\ \footnotesize{[0.7188, 0.8616]}} & \shortstack{0.8068 \\ \footnotesize{[0.7355, 0.8721]}} & \shortstack{\bfseries 0.8163 \\ \footnotesize{[0.7472, 0.8779]}} & \shortstack{0.8212 \\ \footnotesize{[0.7536, 0.8812]}} \\
                              & GPTOSS-20B               & \shortstack{\bfseries 0.2337 \\ \footnotesize{[0.1848, 0.3056]}} & \shortstack{0.2182 \\ \footnotesize{[0.1743, 0.2858]}} & \shortstack{0.2017 \\ \footnotesize{[0.1592, 0.2715]}} & \shortstack{0.2039 \\ \footnotesize{[0.1607, 0.2743]}} & \shortstack{0.2138 \\ \footnotesize{[0.1661, 0.2811]}} & \shortstack{0.2040 \\ \footnotesize{[0.1605, 0.2711]}} & \shortstack{0.2076 \\ \footnotesize{[0.1647, 0.2750]}} \\
                              & Gemma4-31B (aggregate)                & \shortstack{0.2733 \\ \footnotesize{[0.2191, 0.3494]}} & \shortstack{0.2501 \\ \footnotesize{[0.2039, 0.3163]}} & \shortstack{0.2452 \\ \footnotesize{[0.1966, 0.3149]}} & \shortstack{0.2480 \\ \footnotesize{[0.2016, 0.3190]}} & \shortstack{\bfseries 0.2788 \\ \footnotesize{[0.2215, 0.3585]}} & \shortstack{0.2783 \\ \footnotesize{[0.2224, 0.3593]}} & \shortstack{0.2824 \\ \footnotesize{[0.2230, 0.3597]}} \\
                              & Gemma4-31B (stratified, L1-L2) & \shortstack{0.2304 \\ \footnotesize{[0.1128,0.3930]}} & \shortstack{0.2149 \\ \footnotesize{[0.1133,0.4087]}} & \shortstack{0.3065 \\ \footnotesize{[0.1579,0.5485]}} & \shortstack{0.3321 \\ \footnotesize{[0.1684,0.5746]}} & \shortstack{0.2000 \\ \footnotesize{[0.1059,0.3889]}} & \shortstack{0.2000 \\ \footnotesize{[0.1059,0.3889]}}  & \shortstack{0.2824 \\ \footnotesize{[0.2230, 0.3597]}}\\
                              & Gemma4-31B (stratified, L3) & \shortstack{0.3021 \\ \footnotesize{[0.2114,0.4512]}} & \shortstack{0.2668 \\ \footnotesize{[0.1870,0.4034]}} & \shortstack{0.3185 \\ \footnotesize{[0.2098,0.4689]}} & \shortstack{\bfseries 0.3290 \\ \footnotesize{[0.2159,0.4771]}} & \shortstack{0.3080 \\ \footnotesize{[0.2125,0.4559]}} & \shortstack{0.3193 \\ \footnotesize{[0.2167,0.4558]}}  & \shortstack{0.2824 \\ \footnotesize{[0.2230, 0.3597]}}\\
                              & Gemma4-31B (stratified, L4) & \shortstack{0.3021 \\ \footnotesize{[0.2114,0.4512]}} & \shortstack{0.2668 \\ \footnotesize{[0.1870,0.4034]}} & \shortstack{0.3185 \\ \footnotesize{[0.2098,0.4689]}} & \shortstack{\bfseries 0.3290 \\ \footnotesize{[0.2159,0.4771]}} & \shortstack{0.3080 \\ \footnotesize{[0.2125,0.4559]}} & \shortstack{0.3193 \\ \footnotesize{[0.2167,0.4558]}}  & \shortstack{0.2824 \\ \footnotesize{[0.2230, 0.3597]}}\\
                              & Gemma4-31B (stratified, L5) & \shortstack{0.3021 \\ \footnotesize{[0.2114,0.4512]}} & \shortstack{0.2668 \\ \footnotesize{[0.1870,0.4034]}} & \shortstack{0.3185 \\ \footnotesize{[0.2098,0.4689]}} & \shortstack{\bfseries 0.3290 \\ \footnotesize{[0.2159,0.4771]}} & \shortstack{0.3080 \\ \footnotesize{[0.2125,0.4559]}} & \shortstack{0.3193 \\ \footnotesize{[0.2167,0.4558]}}  & \shortstack{0.2824 \\ \footnotesize{[0.2230, 0.3597]}} \\
                              & GPT-4o                   & \shortstack{0.3759 \\ \footnotesize{[0.2945, 0.4770]}} & \shortstack{0.4616 \\ \footnotesize{[0.3657, 0.5691]}} & \shortstack{0.4897 \\ \footnotesize{[0.3908, 0.5974]}} & \shortstack{0.5155 \\ \footnotesize{[0.4144, 0.6186]}} & \shortstack{0.5368 \\ \footnotesize{[0.4326, 0.6413]}} & \shortstack{\bfseries 0.5454 \\ \footnotesize{[0.4418, 0.6486]}} & \shortstack{0.5454 \\ \footnotesize{[0.4418, 0.6486]}} \\
                              & Gemini-2.5Pro            & \shortstack{0.1671 \\ \footnotesize{[0.1098, 0.2816]}} & \shortstack{\bfseries 0.1864 \\ \footnotesize{[0.1201, 0.3144]}} & \shortstack{0.1684 \\ \footnotesize{[0.1108, 0.2762]}} & \shortstack{0.1725 \\ \footnotesize{[0.1118, 0.2887]}} & \shortstack{0.1323 \\ \footnotesize{[0.0905, 0.2056]}} & \shortstack{0.1334 \\ \footnotesize{[0.0912, 0.2080]}} & \shortstack{0.1334 \\ \footnotesize{[0.0912, 0.2080]}} \\
    \midrule
    \multirow{9}{*}{GPQA}    & Qwen3.5-2B               & \shortstack{0.4493 \\ \footnotesize{[0.3463, 0.5975]}} & \shortstack{0.4511 \\                               \footnotesize{[0.3496, 0.5955]}} & \shortstack{0.4977 \\ \footnotesize{[0.3840, 0.6494]}} &                                               \shortstack{\bfseries 0.5211 \\ \footnotesize{[0.3993, 0.6618]}} & \shortstack{0.4323 \\                                                  \footnotesize{[0.3407, 0.5653]}} & \shortstack{0.4522 \\ \footnotesize{[0.3545, 0.5932]}} &                                              \shortstack{0.4383 \\ \footnotesize{[0.3434, 0.5737]}} \\
                              & Qwen3.5-9B               & \shortstack{0.1527 \\ \footnotesize{[0.0934, 0.2839]}} & \shortstack{0.2001 \\ \footnotesize{[0.1242, 0.3439]}} & \shortstack{\bfseries 0.3015 \\ \footnotesize{[0.1513, 0.4854]}} & \shortstack{0.2619 \\ \footnotesize{[0.1285, 0.4430]}} & \shortstack{0.2493 \\ \footnotesize{[0.1284, 0.4196]}} & \shortstack{0.2674 \\ \footnotesize{[0.1382, 0.4493]}} & \shortstack{0.2674 \\ \footnotesize{[0.1382, 0.4493]}} \\
                              & Qwen3.5-122B-A10B        & \shortstack{0.1896 \\ \footnotesize{[0.1199, 0.3219]}} & \shortstack{0.1911 \\ \footnotesize{[0.1219, 0.3229]}} & \shortstack{0.2447 \\ \footnotesize{[0.1423, 0.3977]}} & \shortstack{0.2778 \\ \footnotesize{[0.1701, 0.4594]}} & \shortstack{0.3163 \\ \footnotesize{[0.1952, 0.5089]}} & \shortstack{\bfseries 0.3270 \\ \footnotesize{[0.2021, 0.5253]}} & \shortstack{0.3270 \\ \footnotesize{[0.2021, 0.5253]}} \\
                              & Llama3.1-8B-Instruct     & \shortstack{\bfseries 0.6118 \\ \footnotesize{[0.5229, 0.7220]}} & \shortstack{0.5819 \\ \footnotesize{[0.4954, 0.6917]}} & \shortstack{0.5509 \\ \footnotesize{[0.4700, 0.6553]}} & \shortstack{0.5660 \\ \footnotesize{[0.4813, 0.6724]}} & \shortstack{0.5711 \\ \footnotesize{[0.4858, 0.6800]}} & \shortstack{0.5706 \\ \footnotesize{[0.4848, 0.6801]}} & \shortstack{0.5703 \\ \footnotesize{[0.4845, 0.6799]}} \\
                              & GPTOSS-20B               & \shortstack{0.4265 \\ \footnotesize{[0.3167, 0.5711]}} & \shortstack{0.4651 \\ \footnotesize{[0.3449, 0.6087]}} & \shortstack{0.4611 \\ \footnotesize{[0.3374, 0.6093]}} & \shortstack{0.4423 \\ \footnotesize{[0.3238, 0.5897]}} & \shortstack{0.4413 \\ \footnotesize{[0.3342, 0.5901]}} & \shortstack{\bfseries 0.4685 \\ \footnotesize{[0.3574, 0.6142]}} & \shortstack{0.5168 \\ \footnotesize{[0.3955, 0.6701]}} \\
                              & Gemma4-31B               & \shortstack{0.3774 \\ \footnotesize{[0.2533, 0.5293]}} & \shortstack{0.3922 \\ \footnotesize{[0.2696, 0.5465]}} & \shortstack{0.4165 \\ \footnotesize{[0.2914, 0.5670]}} & \shortstack{\bfseries 0.4465 \\ \footnotesize{[0.3139, 0.5983]}} & \shortstack{0.3733 \\ \footnotesize{[0.2622, 0.5275]}} & \shortstack{0.4030 \\ \footnotesize{[0.2855, 0.5630]}} & \shortstack{0.3974 \\ \footnotesize{[0.2807, 0.5550]}} \\
                              & GPT-4o                   & \shortstack{0.5210 \\ \footnotesize{[0.4299, 0.6358]}} & \shortstack{0.5863 \\ \footnotesize{[0.4881, 0.6926]}} & \shortstack{0.6025 \\ \footnotesize{[0.5060, 0.7199]}} & \shortstack{0.6025 \\ \footnotesize{[0.5058, 0.7231]}} & \shortstack{\bfseries 0.6402 \\ \footnotesize{[0.5399, 0.7545]}} & \shortstack{0.6349 \\ \footnotesize{[0.5354, 0.7505]}} & \shortstack{0.6349 \\ \footnotesize{[0.5354, 0.7505]}} \\
    \midrule
    \multirow{5}{*}{LiveCodeBench}& Qwen3.5-2B               & \shortstack{\bfseries 0.8487 \\ \footnotesize{[0.7951, 0.8980]}} & \shortstack{0.8257 \\ \footnotesize{[0.7623, 0.8898]}} & \shortstack{0.8479 \\ \footnotesize{[0.7897, 0.9011]}} & \shortstack{0.8323 \\ \footnotesize{[0.7719, 0.8923]}} & \shortstack{0.8424 \\ \footnotesize{[0.7855, 0.8951]}} & \shortstack{0.8434 \\ \footnotesize{[0.7853, 0.8966]}} & \shortstack{0.8410 \\ \footnotesize{[0.7827, 0.8955]}} \\
                              & Qwen3.5-122B-A10B        & \shortstack{\bfseries 0.6071 \\ \footnotesize{[0.5379, 0.6792]}} & \shortstack{0.6017 \\ \footnotesize{[0.5339, 0.6729]}} & \shortstack{0.5751 \\ \footnotesize{[0.5094, 0.6462]}} & \shortstack{0.5574 \\ \footnotesize{[0.4922, 0.6332]}} & \shortstack{0.5912 \\ \footnotesize{[0.5229, 0.6653]}} & \shortstack{0.5940 \\ \footnotesize{[0.5256, 0.6682]}} & \shortstack{0.5919 \\ \footnotesize{[0.5241, 0.6671]}} \\
                              & GPTOSS-20B               & \shortstack{0.6587 \\ \footnotesize{[0.5934, 0.7270]}} & \shortstack{\bfseries 0.6686 \\ \footnotesize{[0.6033, 0.7408]}} & \shortstack{0.6276 \\ \footnotesize{[0.5660, 0.7025]}} & \shortstack{0.6489 \\ \footnotesize{[0.5846, 0.7220]}} & \shortstack{0.6472 \\ \footnotesize{[0.5843, 0.7218]}} & \shortstack{0.6409 \\ \footnotesize{[0.5783, 0.7141]}} & \shortstack{0.6361 \\ \footnotesize{[0.5734, 0.7107]}} \\
                              & Gemma4-31B               & \shortstack{0.4201 \\ \footnotesize{[0.3269, 0.5364]}} & \shortstack{\bfseries 0.4297 \\ \footnotesize{[0.3391, 0.5533]}} & \shortstack{0.3827 \\ \footnotesize{[0.3014, 0.4994]}} & \shortstack{0.3807 \\ \footnotesize{[0.3010, 0.4923]}} & \shortstack{0.3742 \\ \footnotesize{[0.2959, 0.4837]}} & \shortstack{0.3741 \\ \footnotesize{[0.2958, 0.4831]}} & \shortstack{0.3741 \\ \footnotesize{[0.2958, 0.4831]}} \\
    \bottomrule
  \end{tabular}
  }
\caption{PR-AUC for different window sizes $T$. Configurations excluded from pooling analysis due to capability floor or ceiling criteria are still included. }
  \label{tab:pr_auc}
\end{table*}





\begin{table*}[htbp]
  \centering

  \resizebox{\textwidth}{!}{%
  \begin{tabular}{@{}
    ll
    S[table-format=2.2(2)]
    S[table-format=2.2(2)]
    S[table-format=2.2(2)]
    S[table-format=2.2(2)]
    S[table-format=2.2(2)]
    S[table-format=2.2(2)]
    S[table-format=2.2(2)]
    @{}
  }
    \toprule
    \multirow{2}{*}{Benchmark} & \multirow{2}{*}{Model} & \multicolumn{6}{c}{PR-AUC}  \\
    \cmidrule(l){3-8}
                              &                      & {$T=128$}             & {$T=256$}       & {$T=400$} & {$T=512$} & {$T=1024$} & {$T=2048$} \\
    \midrule
    \multirow{4}{*}{GSM8K}   & Qwen3.5-2B & \shortstack{-0.0081 \\ \footnotesize{[-0.0587,                             +0.0421]}} &                                             \shortstack{\bfseries +0.0047 \\                                           \footnotesize{[-0.0357, +0.0445]}} &                                    \shortstack{+0.0038 \\                             \footnotesize{[-0.0239, +0.0314]}} & \shortstack{-0.0002 \\                                      \footnotesize{[-0.0243, +0.0236]}} & \shortstack{-0.0003 \\                             \footnotesize{[-0.0160,                                     +0.0160]}} & \shortstack{+0.0020 \\                             \footnotesize{[-0.0093, +0.0142]}} \\
                              & Llama3.1-8B-Instruct     & \shortstack{-0.1463 \\ \footnotesize{[-0.2135, -0.0808]}} & \shortstack{-0.0230 \\ \footnotesize{[-0.0634, +0.0195]}} & \shortstack{+0.0160 \\ \footnotesize{[-0.0109, +0.0478]}} & \shortstack{\bfseries +0.0198 \\ \footnotesize{[-0.0057, +0.0499]}} & \shortstack{-0.0022 \\ \footnotesize{[-0.0062, +0.0009]}} & \shortstack{-0.0011 \\ \footnotesize{[-0.0039, +0.0000]}}\\
                              & GPTOSS-20B                & \shortstack{-0.0980 \\ \footnotesize{[-0.1952, -0.0060]}} & \shortstack{-0.0514 \\ \footnotesize{[-0.1363, +0.0262]}} & \shortstack{-0.0464 \\ \footnotesize{[-0.1261, +0.0259]}} & \shortstack{-0.0541 \\ \footnotesize{[-0.1261, +0.0089]}} & \shortstack{-0.0375 \\ \footnotesize{[-0.0851, -0.0024]}} & \shortstack{\bfseries -0.0166 \\ \footnotesize{[-0.0518, +0.0104]}} \\

    \midrule
    \multirow{17}{*}{MATH500} & Qwen3.5-2B & \shortstack{-0.0731 \\ \footnotesize{[-0.1516,                             -0.0011]}} & \shortstack{-0.0792 \\ \footnotesize{[-0.1542,                             -0.0111]}} & \shortstack{-0.0377 \\ \footnotesize{[-0.1121,                             +0.0285]}} & \shortstack{-0.0134 \\ \footnotesize{[-0.0832,                             +0.0523]}} & \shortstack{\bfseries +0.0149 \\                                           \footnotesize{[-0.0347, +0.0642]}} & \shortstack{-0.0236 \\                             \footnotesize{[-0.0635, +0.0118]}}\\
                              & Qwen3.5-9B & \shortstack{-0.0205 \\ \footnotesize{[-0.0682, +0.0213]}} & \shortstack{+0.0252 \\ \footnotesize{[-0.0130, +0.0700]}} & \shortstack{\bfseries +0.0332 \\ \footnotesize{[-0.0036, +0.0916]}} & \shortstack{+0.0324 \\ \footnotesize{[-0.0010, +0.0824]}} & \shortstack{-0.0002 \\ \footnotesize{[-0.0017, +0.0010]}} & \shortstack{+0.0000 \\ \footnotesize{[+0.0000, +0.0000]}}\\
                              & Qwen3.5-122B-A10B        & \shortstack{-0.0359 \\ \footnotesize{[-0.0861, +0.0020]}} & \shortstack{+0.0089 \\ \footnotesize{[-0.0342, +0.0501]}} & \shortstack{\bfseries +0.0366 \\ \footnotesize{[-0.0071, +0.0884]}} & \shortstack{+0.0221 \\ \footnotesize{[-0.0138, +0.0628]}} & \shortstack{+0.0046 \\ \footnotesize{[-0.0063, +0.0189]}} & \shortstack{+0.0000 \\ \footnotesize{[+0.0000, +0.0000]}} \\
                              & Llama3.1-8B-Instruct     & \shortstack{-0.0595 \\ \footnotesize{[-0.1214, -0.0005]}} & \shortstack{-0.0213 \\ \footnotesize{[-0.0640, +0.0198]}} & \shortstack{-0.0243 \\ \footnotesize{[-0.0594, +0.0080]}} & \shortstack{-0.0268 \\ \footnotesize{[-0.0553, -0.0023]}} & \shortstack{-0.0141 \\ \footnotesize{[-0.0294, -0.0034]}} & \shortstack{\bfseries -0.0048 \\ \footnotesize{[-0.0136, +0.0008]}}\\
                              & GPTOSS-20B               & \shortstack{\bfseries +0.0255 \\ \footnotesize{[-0.0273, +0.0784]}} & \shortstack{+0.0099 \\ \footnotesize{[-0.0367, +0.0522]}} & \shortstack{-0.0052 \\ \footnotesize{[-0.0536, +0.0438]}} & \shortstack{-0.0028 \\ \footnotesize{[-0.0489, +0.0429]}} & \shortstack{+0.0043 \\ \footnotesize{[-0.0420, +0.0499]}} & \shortstack{-0.0034 \\ \footnotesize{[-0.0356, +0.0291]}}\\
                              & Gemma4-31B (aggregate)                & \shortstack{-0.0105 \\ \footnotesize{[-0.0836, +0.0539]}} & \shortstack{-0.0310 \\ \footnotesize{[-0.0896, +0.0271]}} & \shortstack{-0.0357 \\ \footnotesize{[-0.0853, +0.0108]}} & \shortstack{-0.0325 \\ \footnotesize{[-0.0754, +0.0104]}} & \shortstack{\bfseries -0.0009 \\ \footnotesize{[-0.0145, +0.0137]}} & \shortstack{-0.0021 \\ \footnotesize{[-0.0118, +0.0068]}}  \\
                              & Gemma-31B (stratified, L1-L2) & \shortstack{+0.0234 \\ \footnotesize{[-0.1483, +0.2023]}} & \shortstack{+0.0269 \\ \footnotesize{[-0.0802, +0.1501]}} & \shortstack{\bfseries +0.1140 \\ \footnotesize{[+0.0123, +0.2654]}} & \shortstack{+0.1119 \\ \footnotesize{[+0.0176, +0.2515]}} & \shortstack{-0.0001 \\ \footnotesize{[-0.0008, +0.0000]}} & \shortstack{+0.0000 \\ \footnotesize{[+0.0000, +0.0000]}} \\
                              & Gemma4-31B (stratified, L3) & \shortstack{\bfseries +0.0715 \\ \footnotesize{[-0.0846, +0.2270]}} & \shortstack{+0.0270 \\ \footnotesize{[-0.0669, +0.1371]}} & \shortstack{+0.0552 \\ \footnotesize{[-0.0257, +0.1521]}} & \shortstack{+0.0246 \\ \footnotesize{[-0.0066, +0.0727]}} & \shortstack{-0.0126 \\ \footnotesize{[-0.0524, +0.0019]}} & \shortstack{+0.0000 \\ \footnotesize{[+0.0000, +0.0000]}} \\
                              & Gemma4-31B (stratified, L4) & \shortstack{\bfseries +0.0849 \\ \footnotesize{[-0.0561, +0.2317]}} & \shortstack{+0.0748 \\ \footnotesize{[-0.0018, +0.1676]}} & \shortstack{+0.0068 \\ \footnotesize{[-0.0566, +0.0795]}} & \shortstack{+0.0055 \\ \footnotesize{[-0.0488, +0.0607]}} & \shortstack{+0.0130 \\ \footnotesize{[-0.0024, +0.0337]}} & \shortstack{+0.0022 \\ \footnotesize{[-0.0044, +0.0100]}} \\
                              & Gemma4-31B (stratified, L5) & \shortstack{+0.0291 \\ \footnotesize{[-0.1073, +0.1704]}} & \shortstack{-0.0326 \\ \footnotesize{[-0.1466, +0.0805]}} & \shortstack{+0.0589 \\ \footnotesize{[-0.0431, +0.1731]}} & \shortstack{\bfseries +0.0651 \\ \footnotesize{[-0.0260, +0.1593]}} & \shortstack{-0.0287 \\ \footnotesize{[-0.0636, -0.0022]}} & \shortstack{-0.0208 \\ \footnotesize{[-0.0488, -0.0026]}} \\
                              & GPT-4o                   & \shortstack{-0.1661 \\ \footnotesize{[-0.2534, -0.0814]}} & \shortstack{-0.0821 \\ \footnotesize{[-0.1571, -0.0056]}} & \shortstack{-0.0550 \\ \footnotesize{[-0.1221, +0.0109]}} & \shortstack{-0.0299 \\ \footnotesize{[-0.0749, +0.0157]}} & \shortstack{-0.0084 \\ \footnotesize{[-0.0247, +0.0028]}} & \shortstack{\bfseries +0.0000 \\ \footnotesize{[+0.0000, +0.0000]}}\\
                              & Gemini-2.5Pro            & \shortstack{+0.0406 \\ \footnotesize{[-0.0248, +0.1263]}} & \shortstack{\bfseries +0.0602 \\ \footnotesize{[+0.0155, +0.1323]}} & \shortstack{+0.0394 \\ \footnotesize{[+0.0084, +0.0881]}} & \shortstack{+0.0450 \\ \footnotesize{[+0.0127, +0.1016]}} & \shortstack{-0.0012 \\ \footnotesize{[-0.0051, +0.0019]}} & \shortstack{+0.0000 \\ \footnotesize{[+0.0000, +0.0000]}} \\
    \midrule
    \multirow{10}{*}{GPQA}    & Qwen3.5-2B & \shortstack{+0.0131 \\ \footnotesize{[-0.0940, +0.1206]}} & \shortstack{+0.0146 \\ \footnotesize{[-0.0917, +0.1196]}} & \shortstack{+0.0615 \\ \footnotesize{[-0.0484,+0.1767]}} & \shortstack{\bfseries +0.0777 \\                                           \footnotesize{[-0.0338, +0.1897]}} & \shortstack{-0.0072 \\ \footnotesize{[-0.0871, +0.0675]}} & \shortstack{+0.0150 \\                             \footnotesize{[-0.0395, +0.0732]}}\\
                              & Qwen3.5-9B   & \shortstack{-0.1122 \\ \footnotesize{[-0.2702, +0.0327]}} & \shortstack{-0.0637 \\ \footnotesize{[-0.2120, +0.0708]}} & \shortstack{+0.0290 \\ \footnotesize{[-0.0931, +0.1562]}} & \shortstack{-0.0066 \\ \footnotesize{[-0.1092, +0.1059]}} & \shortstack{-0.0195 \\ \footnotesize{[-0.0487, -0.0030]}} & \shortstack{+0.0000 \\ \footnotesize{[+0.0000, +0.0000]}} \\
                              & Qwen3.5-122B-A10B        & \shortstack{-0.1444 \\ \footnotesize{[-0.3198, +0.0195]}} & \shortstack{-0.1435 \\ \footnotesize{[-0.3068, +0.0017]}} & \shortstack{-0.0957 \\ \footnotesize{[-0.2319, +0.0158]}} & \shortstack{-0.0511 \\ \footnotesize{[-0.1730, +0.0590]}} & \shortstack{-0.0111 \\ \footnotesize{[-0.0328, +0.0031]}} & \shortstack{-0.0000 \\ \footnotesize{[+0.0000, +0.0000]}} \\
                              & Llama3.1-8B-Instruct     & \shortstack{\bfseries +0.0411 \\ \footnotesize{[-0.0354, +0.1191]}} & \shortstack{+0.0111 \\ \footnotesize{[-0.0587, +0.0807]}} & \shortstack{-0.0195 \\ \footnotesize{[-0.0780, +0.0378]}} & \shortstack{-0.0046 \\ \footnotesize{[-0.0507, +0.0393]}} & \shortstack{+0.0009 \\ \footnotesize{[-0.0081, +0.0079]}} & \shortstack{+0.0002 \\ \footnotesize{[-0.0009, +0.0012]}}\\
                              & GPTOSS-20B               & \shortstack{-0.0898 \\ \footnotesize{[-0.2069, +0.0243]}} & \shortstack{-0.0533 \\ \footnotesize{[-0.1609, +0.0540]}} & \shortstack{-0.0562 \\ \footnotesize{[-0.1520, +0.0384]}} & \shortstack{-0.0749 \\ \footnotesize{[-0.1618, +0.0095]}} & \shortstack{-0.0716 \\ \footnotesize{[-0.1449, +0.0012]}} & \shortstack{\bfseries -0.0483 \\ \footnotesize{[-0.1094, +0.0048]}}\\
                              & Gemma4-31B & \shortstack{-0.0243 \\ \footnotesize{[-0.1838, +0.1348]}} & \shortstack{-0.0099 \\ \footnotesize{[-0.1449, +0.1295]}} & \shortstack{+0.0099 \\ \footnotesize{[-0.1162, +0.1340]}} & \shortstack{\bfseries +0.0413 \\ \footnotesize{[-0.0556, +0.1428]}} & \shortstack{-0.0243 \\ \footnotesize{[-0.0552, +0.0003]}} & \shortstack{+0.0055 \\ \footnotesize{[-0.0056, +0.0196]}}\\
                              & GPT-4o   & \shortstack{-0.1126 \\ \footnotesize{[-0.2127, -0.0163]}} & \shortstack{-0.0532 \\ \footnotesize{[-0.1390, +0.0293]}} & \shortstack{-0.0318 \\ \footnotesize{[-0.0888, +0.0233]}} & \shortstack{-0.0314 \\ \footnotesize{[-0.0853, +0.0172]}} & \shortstack{\bfseries +0.0052 \\ \footnotesize{[-0.0036, +0.0150]}} & \shortstack{+0.0000 \\ \footnotesize{[+0.0000, +0.0000]}}\\
    \midrule
    \multirow{5}{*}{LiveCodeBench}& Qwen3.5-2B                & \shortstack{\bfseries +0.0079 \\ \footnotesize{[-0.0497, +0.0648]}} &              \shortstack{-0.0128 \\ \footnotesize{[-0.0727, +0.0464]}} & \shortstack{+0.0076 \\ \footnotesize{[-0.0411, +0.0527]}} & \shortstack{-0.0072 \\ \footnotesize{[-0.0580, +0.0405]}} & \shortstack{+0.0014 \\ \footnotesize{[-0.0312, +0.0295]}} & \shortstack{+0.0024 \\ \footnotesize{[-0.0080, +0.0114]}}\\
                              & Qwen3.5-122B-A10B        & \shortstack{\bfseries +0.0155 \\ \footnotesize{[-0.0511, +0.0823]}} & \shortstack{+0.0093 \\ \footnotesize{[-0.0498, +0.0672]}} & \shortstack{-0.0167 \\ \footnotesize{[-0.0667, +0.0337]}} & \shortstack{-0.0325 \\ \footnotesize{[-0.0780, +0.0149]}} & \shortstack{-0.0006 \\ \footnotesize{[-0.0419, +0.0404]}} & \shortstack{+0.0022 \\ \footnotesize{[-0.0343, +0.0395]}}\\
                              & GPTOSS-20B               & \shortstack{+0.0200 \\ \footnotesize{[-0.0288, +0.0666]}} & \shortstack{\bfseries +0.0312 \\ \footnotesize{[-0.0141, +0.0768]}} & \shortstack{-0.0084 \\ \footnotesize{[-0.0439, +0.0261]}} & \shortstack{+0.0125 \\ \footnotesize{[-0.0219, +0.0483]}} & \shortstack{+0.0112 \\ \footnotesize{[-0.0145, +0.0375]}} & \shortstack{+0.0048 \\ \footnotesize{[-0.0105, +0.0209]}}\\
                              & Gemma4-31B               & \shortstack{+0.0449 \\ \footnotesize{[-0.0343, +0.1263]}} & \shortstack{\bfseries +0.0561 \\ \footnotesize{[+0.0147, +0.1049]}} & \shortstack{+0.0097 \\ \footnotesize{[-0.0094, +0.0323]}} & \shortstack{+0.0069 \\ \footnotesize{[-0.0030, +0.0196]}} & \shortstack{+0.0002 \\ \footnotesize{[-0.0013, +0.0015]}} & \shortstack{+0.0000 \\ \footnotesize{[+0.0000, +0.0000]}}\\
    \bottomrule
  \end{tabular}
  }
    \caption{Delta PR-AUC for different window sizes $T$. Configurations excluded from pooling analysis due to capability floor or ceiling criteria are still included. }
  \label{tab:delta_pr_auc}
\end{table*}

\end{document}